\documentclass[runningheads]{llncs}
\usepackage{geometry}
\usepackage[T1]{fontenc}
\usepackage{rotating}
\usepackage[export]{adjustbox}
\usepackage{graphicx}
\usepackage{amsmath}
\usepackage{subfig}
\usepackage{comment}
\usepackage{gensymb} 
\usepackage{algorithm}
\usepackage{algorithmic}
\usepackage{amssymb}
\usepackage{amsmath}
\usepackage{colortbl}
\usepackage{xcolor}
\usepackage{blindtext}

\begin{document}
\title{Walk the Lines 2: Contour Tracking for Detailed Segmentation of Infrared Ships and Other Objects}

\titlerunning{Walk the Lines 2: Contour Tracking for Detailed Segmentation}
%

\author{Andr\'{e} Peter Kelm\orcidID{0000-0003-4146-7953} \and
Max Braeschke\thanks{Max Breaschke contributed to this work as part of his bachelor thesis and developed the WtL2 binarization method.} \and Emre Gülsoylu\orcidID{0000-0002-3834-3645} \and Simone Frintrop\orcidID{0000-0002-9475-3593}
}

\authorrunning{A. Kelm et al.}
%
\institute{University of Hamburg, Mittelweg 177, 20148 Hamburg, Germany
\email{\{andre.kelm,emre.guelsoylu,simone.frintrop\}@uni-hamburg.de}}

%
\maketitle              
\begin{abstract}
This paper presents Walk the Lines 2 (WtL2), a unique contour tracking algorithm specifically adapted for detailed segmentation of infrared (IR) ships and various objects in RGB.\footnote{Parts of this work will also appear in the forthcoming doctoral dissertation of André Kelm at Helmut Schmidt University \cite{Kelm2025}.} This extends the original Walk the Lines (WtL) \cite{9412410}, which focused solely on detailed ship segmentation in color.
These innovative WtLs can replace the standard non-maximum suppression (NMS) by using contour tracking to refine the object contour until a 1-pixel-wide closed shape can be binarized, forming a segmentable area in foreground-background scenarios.
WtL2 broadens the application range of WtL beyond its original scope, adapting to IR and expanding to diverse objects within the RGB context.
To achieve IR segmentation, we adapt its input, the object contour detector, to IR ships. 
In addition, the algorithm is enhanced to process a wide range of RGB objects, outperforming the latest generation of contour-based methods when achieving a closed object contour, offering high peak Intersection over Union (IoU) with impressive details.
This positions WtL2 as a compelling method for specialized applications that require detailed segmentation or high-quality samples, potentially accelerating progress in several niche areas of image segmentation.
\end{abstract}

\section{Introduction}
Image segmentation is a fundamental task in computer vision, with applications in medical imaging, robotics, autonomous vehicles, agriculture, and the maritime sector \cite{survey,landmarks}; for example, it plays an important role in maritime navigation \cite{navigation} and traffic safety \cite{CHEN2024}.
Fundamental models such as SAM are already powerful in the RGB domain \cite{SAM} 
and are well suited for ship segmentation since ships are a common category in many benchmarks.

In practice, however, there are many niches where fundamental models do not perform optimally out of the box and require customization.
For example, the infrared (IR) domain is widely used in maritime applications, due to its ability to enhance visibility in the maritime environment \cite{visapp18}, but there are few methods or datasets available for IR ship segmentation \cite{noinfosshipsegm}.
Challenges increase when applications require a high level of detail \cite{sam_hq}, which is not uncommon in the maritime and other specialized fields.

To address this, we present ‘Walk the Lines 2’ (WtL2), a method that complements traditional IR ship segmentation techniques, extracts impressive details, and appears to be even more versatile by extending its application to various objects for color images.
Building on the original 
Walk the Lines (WtL) 
\cite{9412410}, which was developed for RGB ship segmentation only, to capture the fine details of ship antennas and superstructures, WtL2 extends these capabilities. 
The adaptation for the IR domain consists of simply replacing its input, the object contour detector designed for color images, with one trained for IR ships.
\begin{figure*}[b!]
\centering
\subfloat[Ori-\\ginal \cite{PixabayImage}] {\includegraphics[width=0.687in]{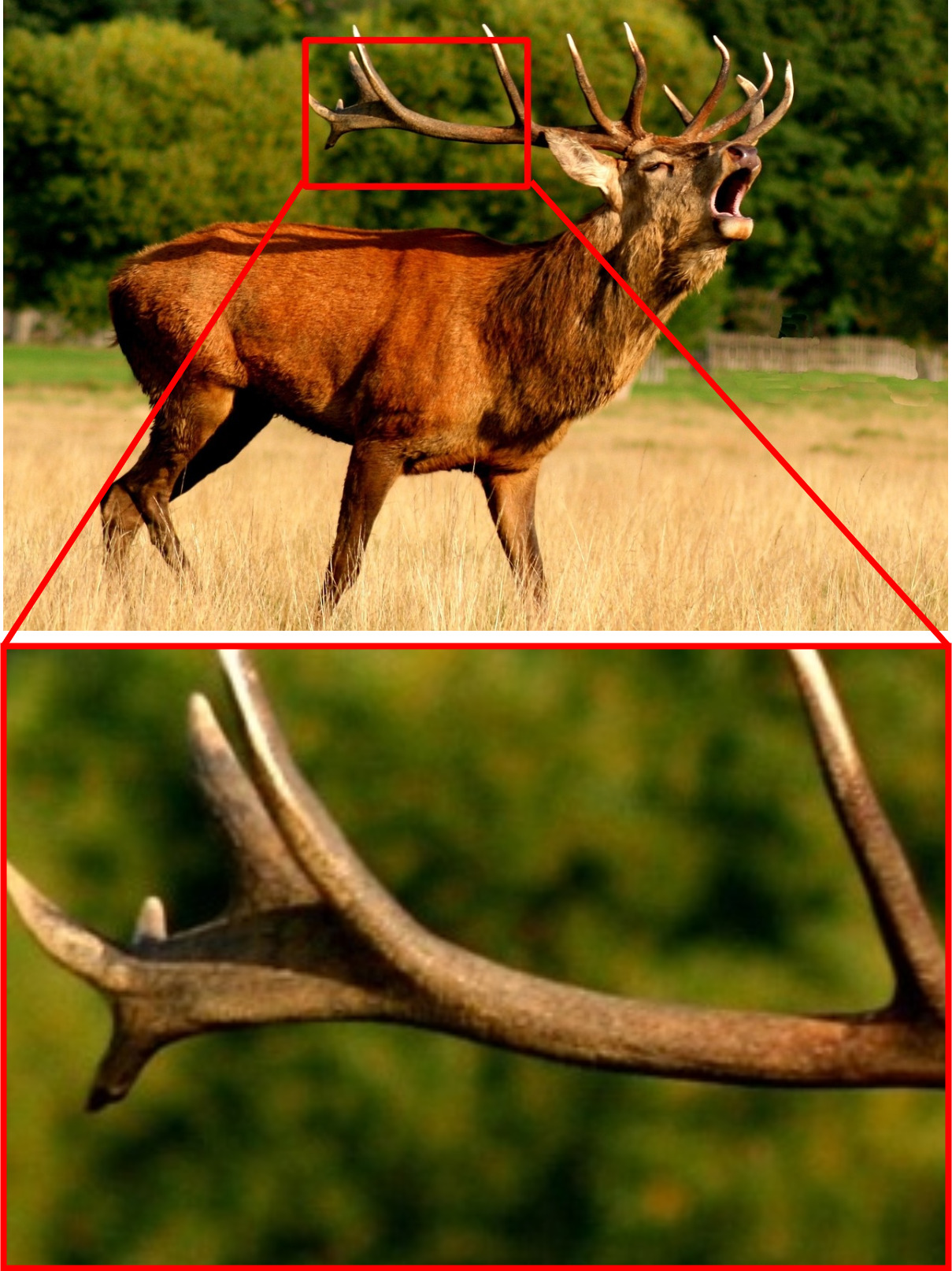}%
\label{orimoto}}
\hfil
\subfloat[soft\\ contour \cite{rcn}]{\includegraphics[width=0.675in]{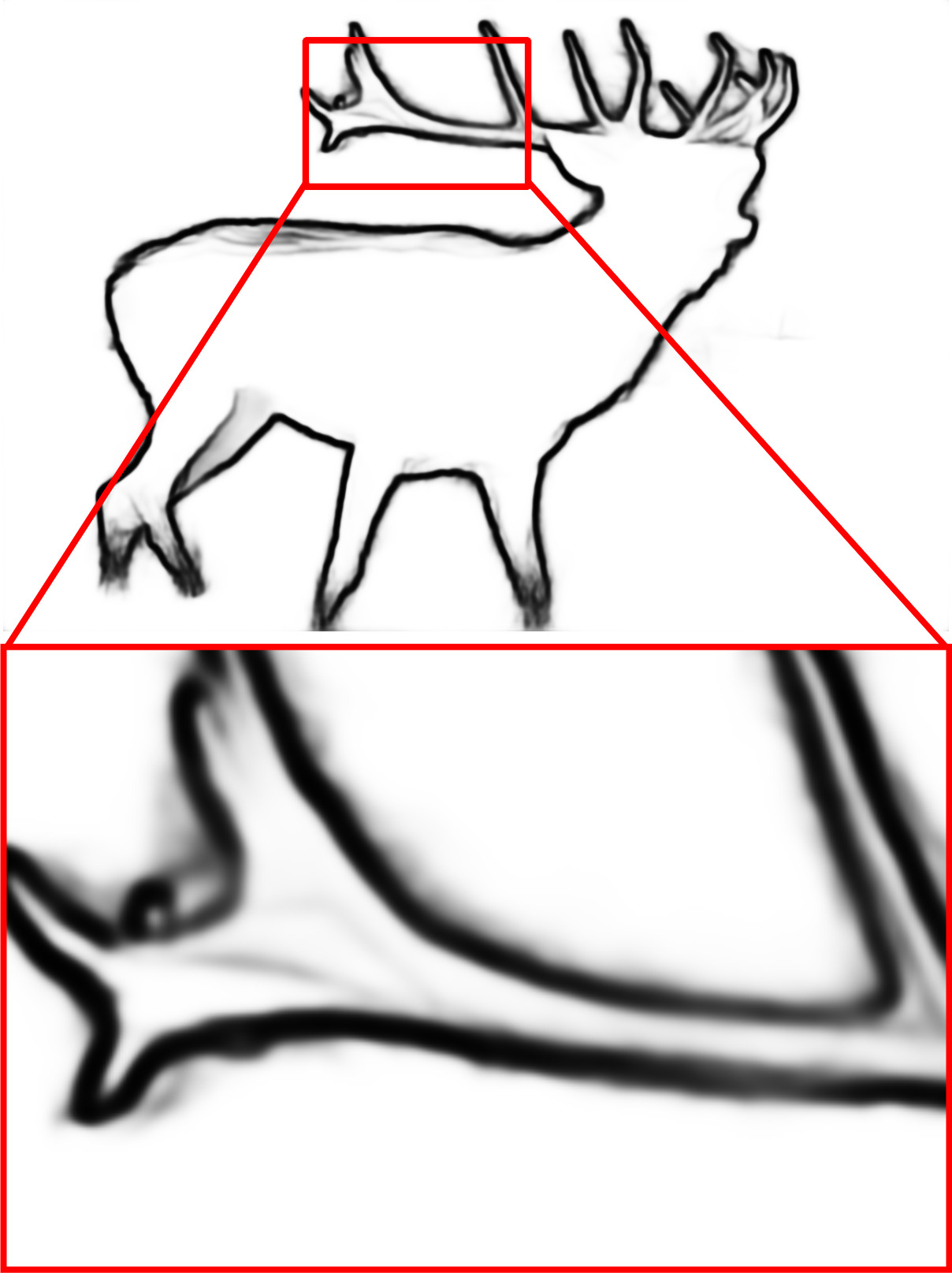}
\label{soft}}
\hfil
\subfloat[NMS \\ \cite{canny1986computational}]{\includegraphics[width=0.675in]{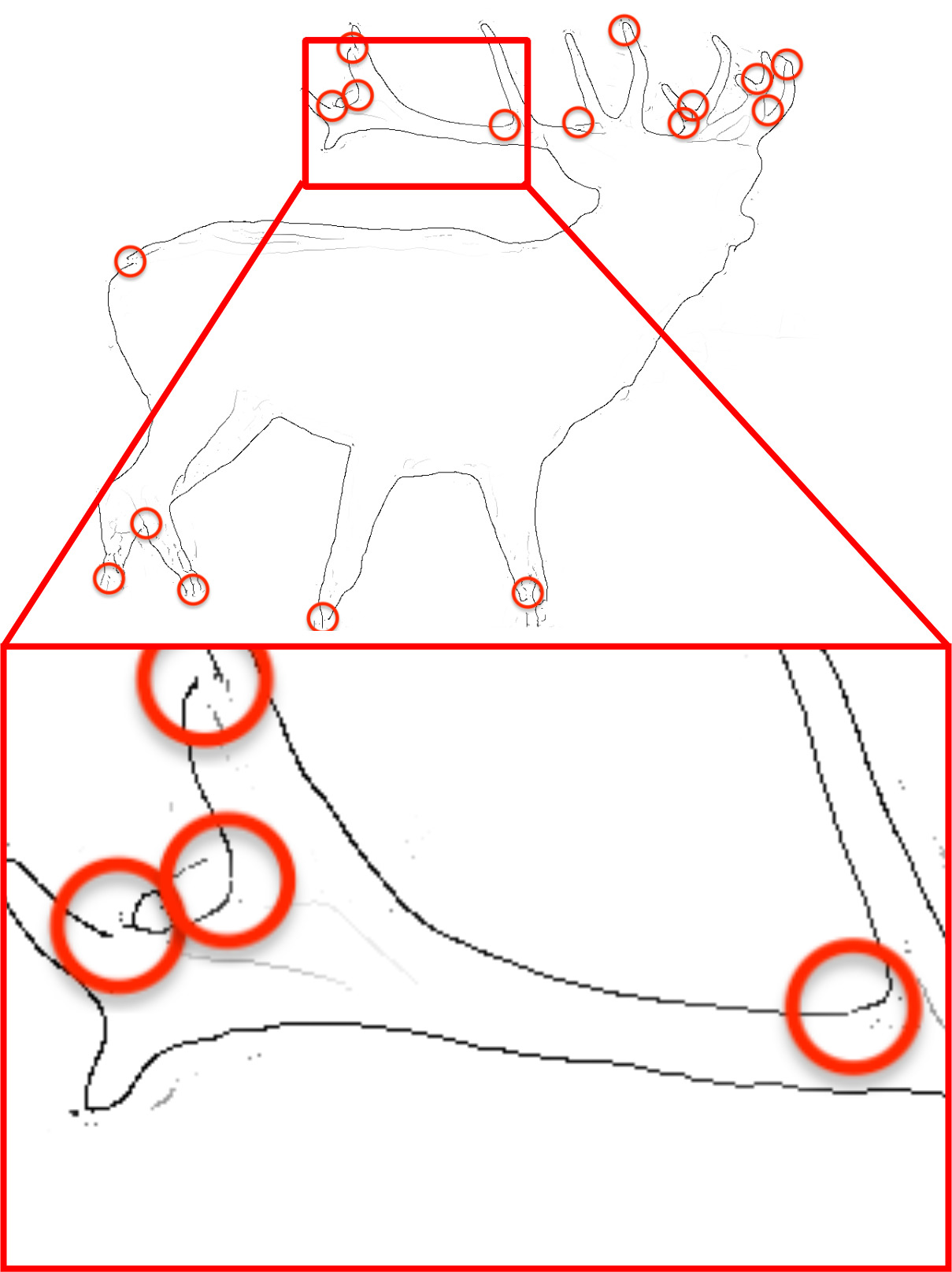}%
\label{nmsdeer}}
\hfil
\subfloat[WtL2-\\contour]{\includegraphics[width=0.675in]{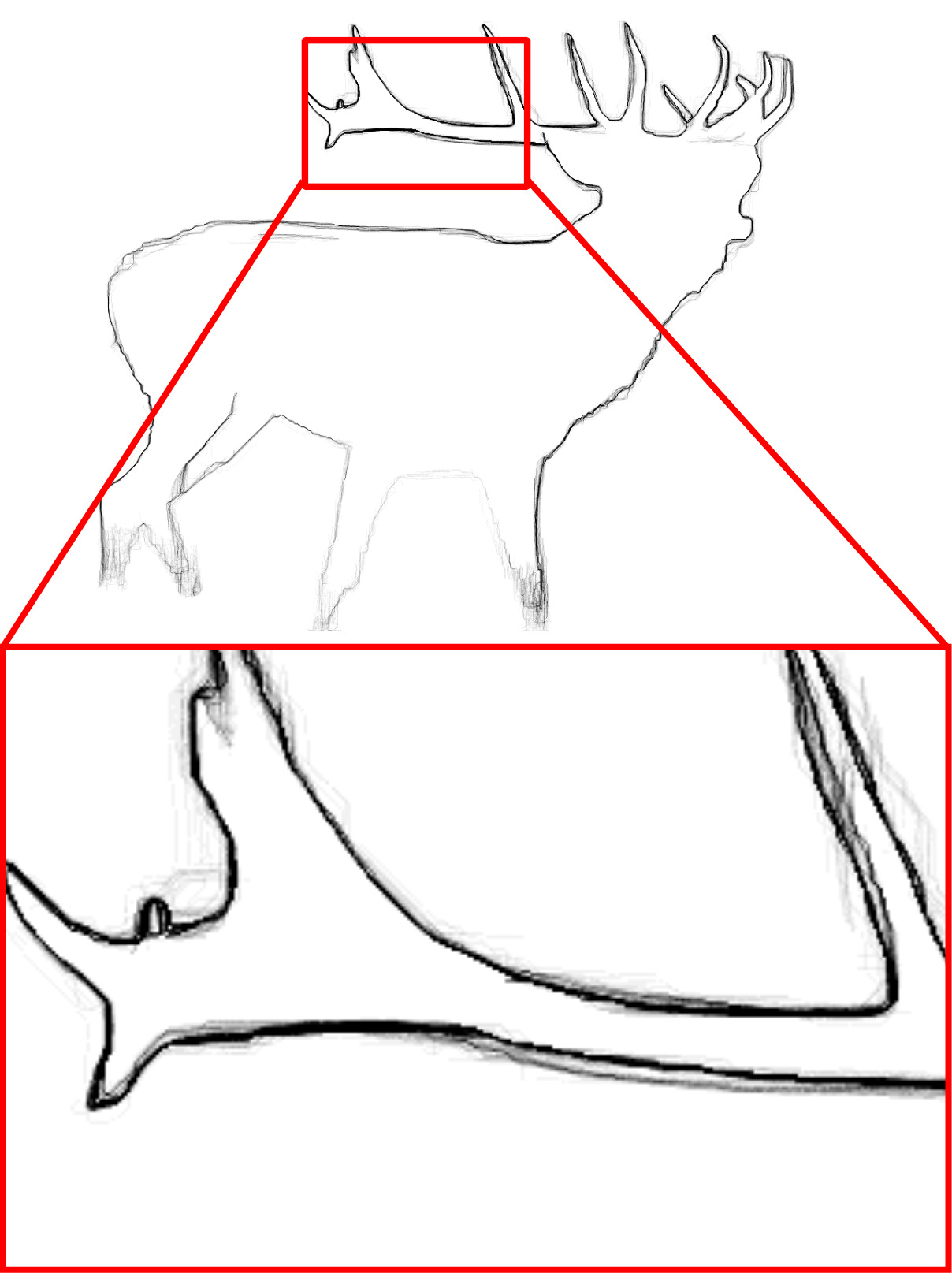}%
\label{wtlcontour}}
\hfil
\subfloat[binary\\WtL2]{\includegraphics[width=0.675in]{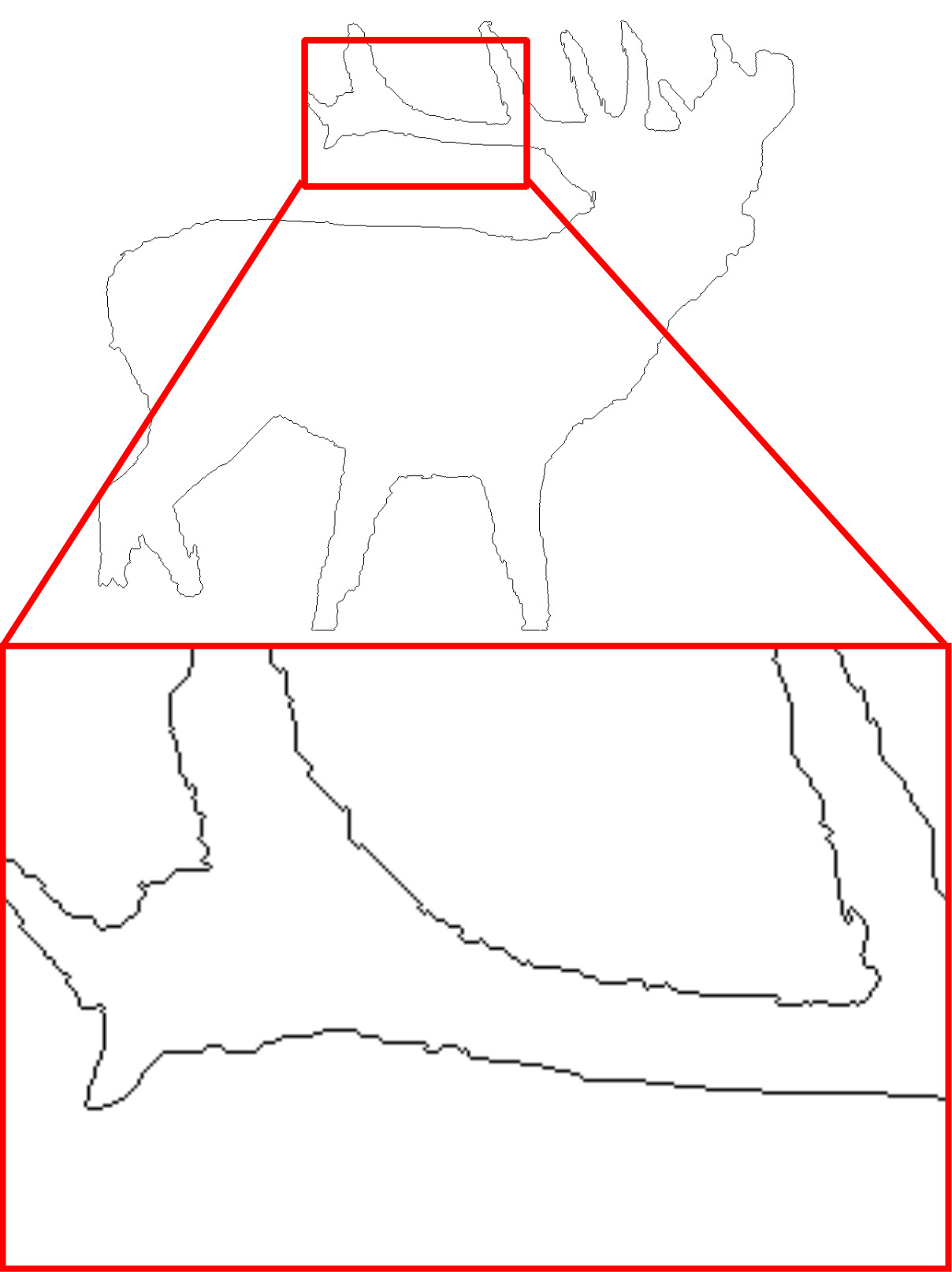}%
\label{wtl2bin}}
\hfil
\subfloat[E2EC\\(sota) \cite{Zhang_2022_CVPR}]{\includegraphics[width=0.675in]{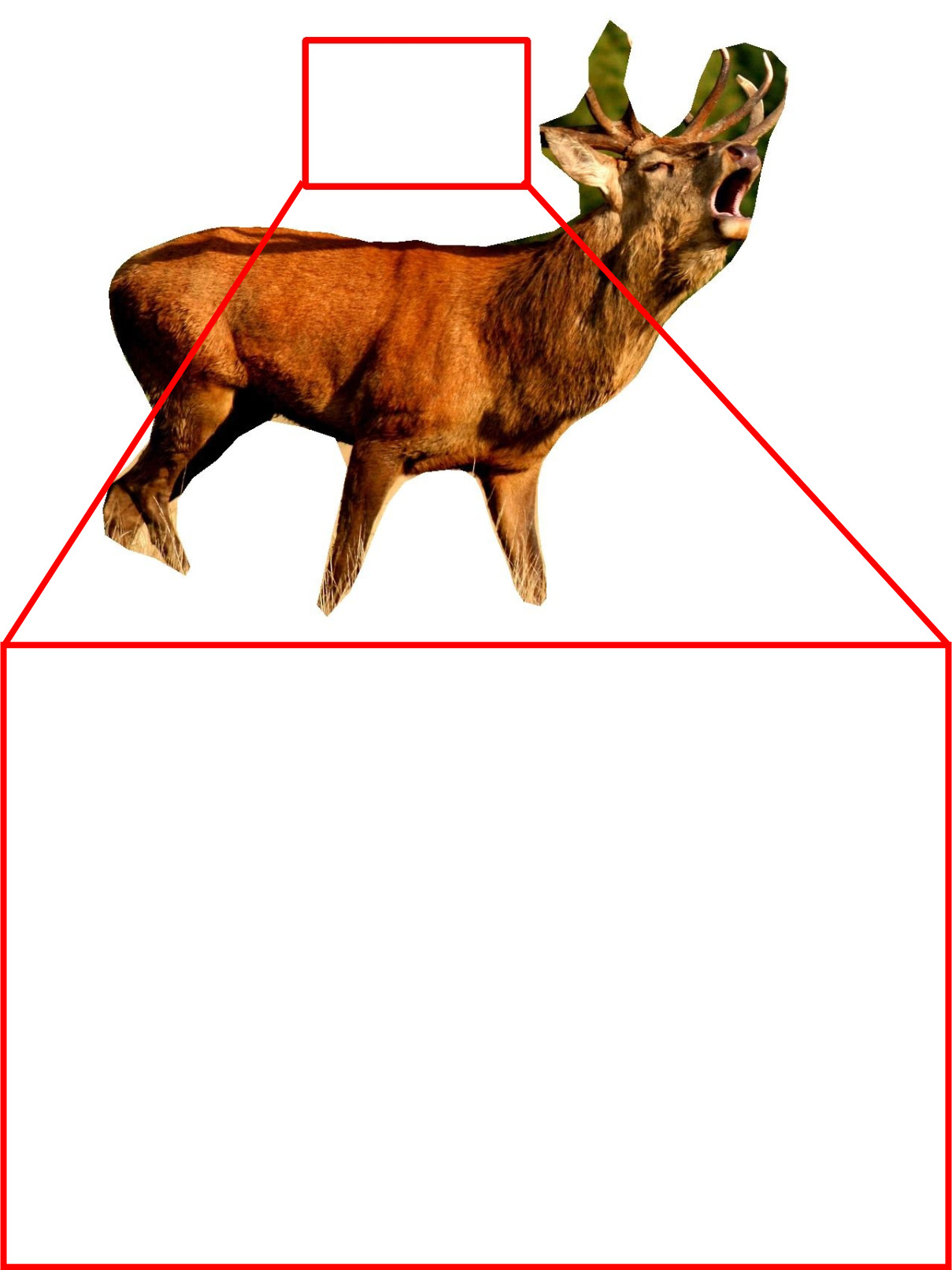}%
\label{sotasegm}}
\hfil
\subfloat[WtL2-\\seg (ours)]{\includegraphics[width=0.675in]{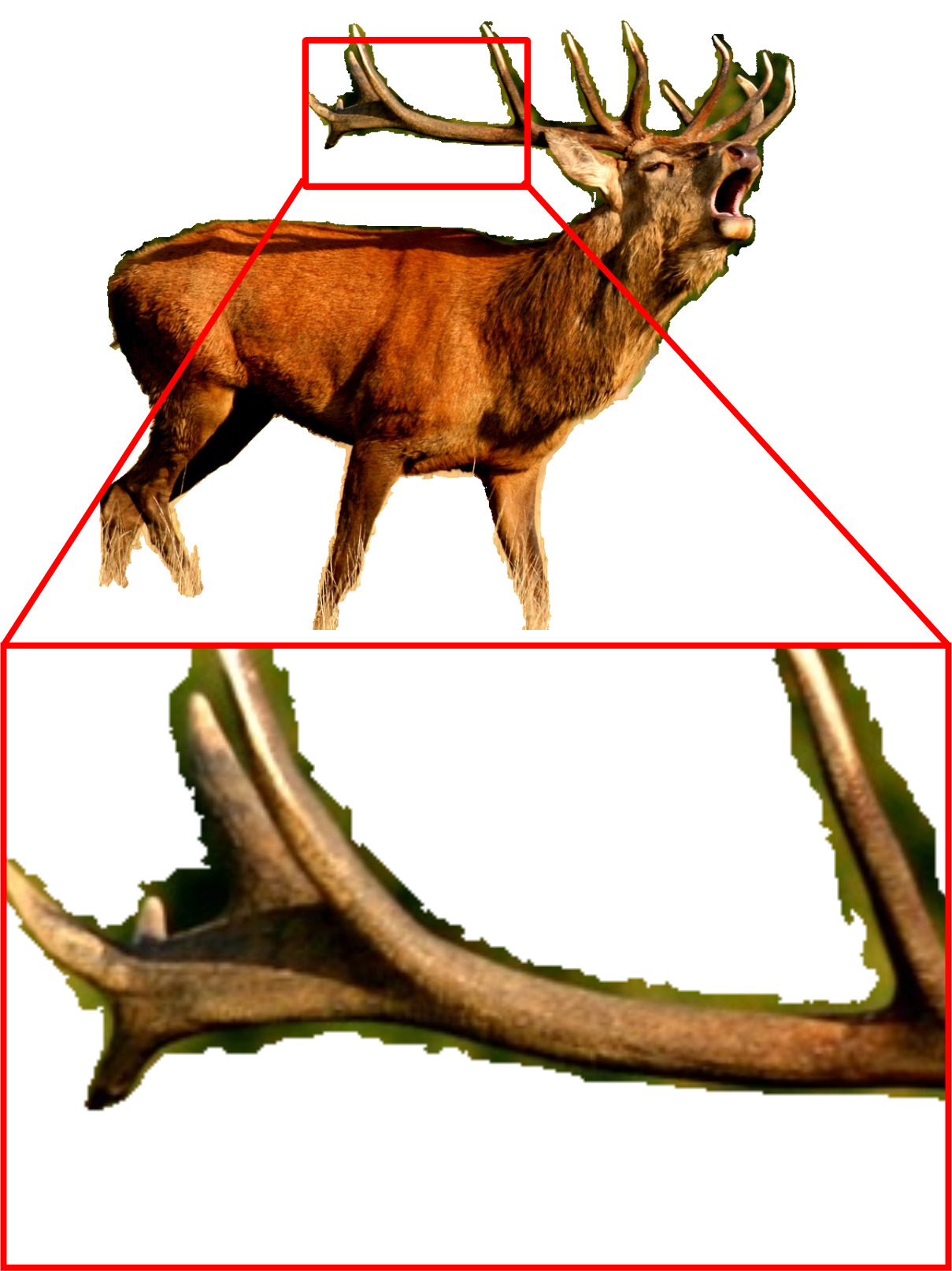}%
\label{wtl2segm}}
\caption{Visualization of \protect\subref{orimoto} the original, which, when processed with an object contour detector
, produces a \protect\subref{soft} soft contour with two post-processing options: \protect\subref{nmsdeer} NMS, which inserts gaps in the contours (marked by red circles), or \protect\subref{wtlcontour} WtL2-contour, which refines the contours and keeps them connected.
Further processing results in the \protect\subref{wtl2bin} 'perfect' binary WtL2. 
We visually compare contour-based segmentation, in particular \protect\subref{sotasegm} E2EC with our \protect\subref{wtl2segm} WtL2-seg.}
\label{explainwtl2}
\end{figure*}

Contour-based segmentation like the WtLs promises detailed results as it is usually based on a contour detection method, providing detailed predictions (Fig.\,\ref{soft}). Non-max\-imum suppression (NMS) is often used for thinning or refinement, but often introduces new gaps in the contour (Fig.\,\ref{nmsdeer}).
To avoid this, 
WtLs use contour tracking to refine object contours along with details (Fig.\,\ref{wtlcontour}), generating closed binary contours (Fig.\,\ref{wtl2bin}).
Other contour-based methods exist: A graphical network can compute the shape of the object's contour \cite{Ling2019graphgcn}, DeepSnake \cite{Peng_2020_CVPR} integrates a learnable active contour into a continuous deep network for better refinement, and E2EC \cite{Zhang_2022_CVPR} further innovates by a learnable contour initialization architecture, multi-direction alignment, and dynamic matching loss, achieving state-of-the-art (sota) for this type of methods. However, our algorithm segments many details better (Fig.\,\ref{sotasegm} vs. \ref{wtl2segm}).
This paper contains two contributions:
\begin{itemize}
    \item We demonstrate the applicability of WtL2 for detailed segmentation of IR ship images. 
    \item Our WtL2 is capable of segmenting a variety of objects, including, but not limited to, cars, dogs, and deer.  
    We have almost doubled the IoU from WtL to WtL2,
    and outperform the sota in contour-based segmentation for images with successfully closed object contours.    
\end{itemize}
While we do not claim that our method consistently outperforms existing benchmarks in terms of overall IoU, it shines in subsets where the algorithm works as intended and a closed shape is recognized. 
WtL2 achieves high IoU peaks and high-quality segmentations even for complex shapes.
This demonstrates the potential of WtL2 for detailed segmentation in a wide range of applications, such as rare object categories or niche areas like IR ship segmentation, as well as other 
specialized fields not yet fully explored. This ability to address niches makes WtL2 a valuable tool for pushing the boundaries of current computer vision applications.

\section{Related Work}
In this section, we first review recent advances in IR ship segmentation. We then discuss the current state in contour-based segmentation, closing 
with a little description to show how our method differs.

\subsection{Infrared Ship Segmentation}
IR ship segmentation presents unique challenges compared to RGB segmentation due to the scarcity of datasets and the limited number of studies and methods, where model weights are often not publicly available \cite{noinfosshipsegm}.

Progress in IR ship segmentation has been hindered by the unavailability of public datasets.
MassMIND \cite{doi:10.1177/02783649231153020} provides a significant contribution with more than 2,900 segmented IR images of coastal areas and seven classes (such as sky, water, bridge, obstacle, living obstacle, etc.), but none explicitly for ships. Unfortunately, this and the presence of the coast do not fit with our foreground/background approach, where the ship and its details are more in focus. 

Some methods use adversarial domain adaptation \cite{noinfosshipsegm} to overcome the lack of IR ship segmentation data.
Other works also try to overcome data scarcity by proposing weakly supervised or semi-supervised methods \cite{weakly}.
The use of foundational models such as SAM for similar purposes is still an emerging area of research \cite{IRSAM}, but also requires annotations \cite{chen2023learningsegmentanythingthermal}. 
In short, despite its critical importance to maritime applications, the maritime IR domain has been underexplored. This highlights the importance of our work in extracting highly detailed ship segmentations from IR imagery.

\subsection{Contour-Based Segmentation of Various Objects}
Contour-based segmentation was a competitive alternative to traditional mask-based segmentation \cite{hierarchicalsegm}, especially before the era of deep learning (DL). 
Despite their incredible performance gains, DL methods initially suffered from a rather blob-like mask, and even today some details are difficult to segment \cite{ZHANG20221}. 
Recent contour-based methods aim to extract even more detail \cite{9157078}, for example by combining advanced object detection with active contours \cite{shipcontour} or embedding them in an end-to-end network, such as Deep Snake \cite{Peng_2020_CVPR}.
E2EC builds on this and shows sota results for contour-based methods using a semantic edge detector as the basis for the active contours \cite{Zhang_2022_CVPR}. 
Our approach differs in that we use an object contour detector whose prediction is not improved by an active contour, but innovatively by a tracking CNN that circles the contour around the object.

\section{Walk the Lines 2}
\begin{figure}[b!]
\centering
\includegraphics[width=1\textwidth]{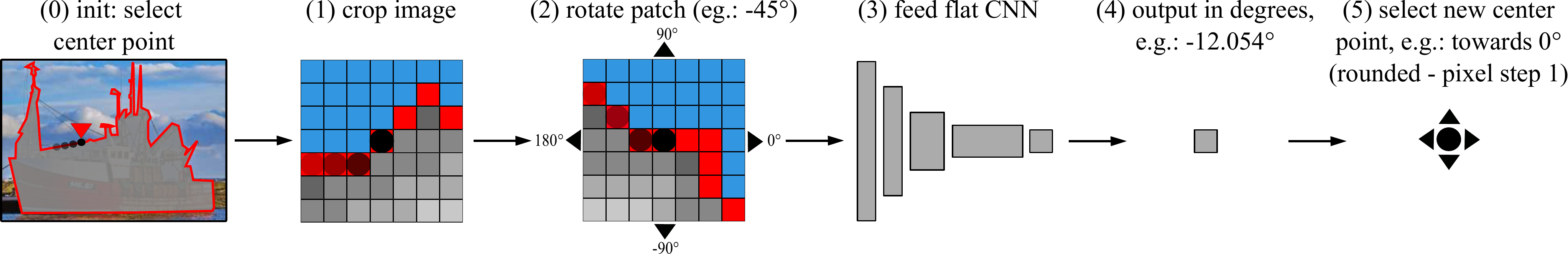}
\caption{Diagram illustrating the contour tracking process of WtLs.}
\label{fig3}
\end{figure}

The \textbf{W}alk \textbf{t}he \textbf{L}ines (WtL) algorithm 
has a very narrow application: detailed segmentation of RGB ships, including antennas and superstructures.
WtL2\footnote{https://github.com/AndreKelm/WalktheLines2} demonstrates that the algorithm is more versatile. It introduces two key contributions:
adapting the algorithm for IR ship segmentation and extending its application to various objects in the RGB domain.
First, we explain the main function of both WtLs: contour tracking. Then we show the process flow, and discuss the implemented contributions in sections \ref{ISSaOC} and \ref{CBSoO}\\
The uniqueness of these algorithms is defined by its contour tracking capability.
A detailed description of this feature and the flat CNN utilized can be found in the original source \cite{9412410}, as the principle of both WtLs remain consistent. The processes
require an object contour detector (we use RefineContourNet (RCN) \cite{rcn}) and are outlined here as well in Fig.~\ref{fig3}:
\textbf{(0) init: select center point:} 
randomly select from ground truth (training) or high confidence value from NMS processed soft object contour map (inference)
\textbf{(1) crop image patch:} centered around the selected point.
\textbf{(2) rotate patch:} based on directional changes retrieved from the previous center point.
\textbf{(3) run tracking CNN:} input is always a $7\times 7\times 4$ patch, concatenating image and soft contour.
\textbf{(4) output in degrees:} example output is $-12.054\degree$. 
\textbf{(5) select new center point:} determine by rounding based on pixel step (stochastically chosen, with a preference for 1 over 2 or 3); for a pixel step of $1$, round to $-135\degree$ (far right), $-90\degree$ (right), $-45\degree$ (slight right), $0\degree$ (straight), $45\degree$ (slight left), $90\degree$ (left), $135\degree$ (far left), or in rare cases, $\pm180\degree$ (turn around).
The process has been scaled with
hundreds of trackers, or thousands depending on the contour, as some have reached dead ends.
The pixel-by-pixel contour tracking of each image, with all trackers running simultaneously, typically takes a few tens of seconds to complete.

\begin{figure}[t!]
\centering
\includegraphics[width=1\textwidth]{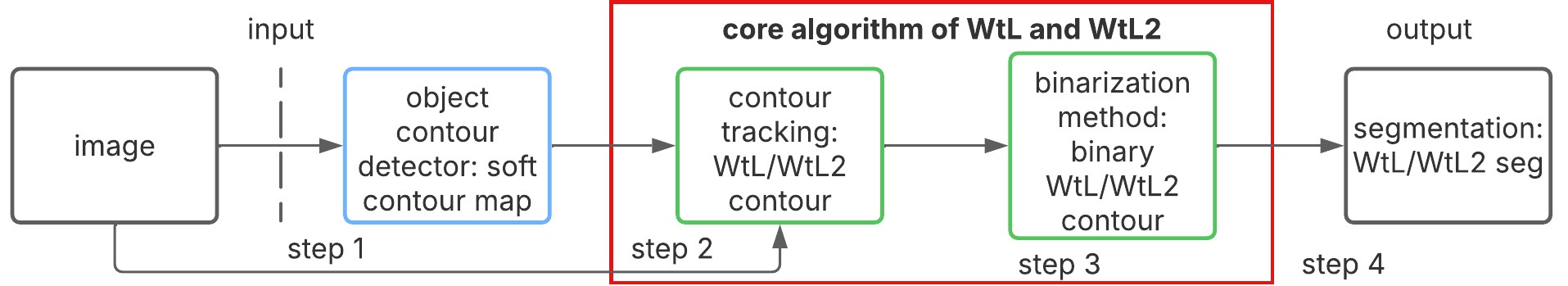}
\caption{Overview of WtL and WtL2 processes. 
Red outlines core algorithm.
Modifications for WtL2 are outlined in color:  adaptation for IR ship via object contour detector retraining in blue, and extension for various objects in green.}
\label{fig1}
\end{figure}

To contextualize contour tracking and highlight our main contributions Fig.~\ref{fig1} outlines the high-level processes that apply to both WtLs. The changes we made are color-coded for clarity.
It uses the results from an object contour detector \cite{rcn} to produce a soft contour (ex. in Fig.~\ref{soft}). For IR ships, it was sufficient to retrain the detector, which allows detailed segmentation in this domain. The already explained unique contour tracking refines its result concatenated with the image (ex. in Fig.~\ref{wtlcontour}).
Further processing results in a closed shape (ex. in Fig.~\ref{wtl2bin}) with a 1-pixel-wide contour that encloses an area that can also be used as a mask (ex. in Fig.\,\ref{wtl2bin}).
To extend to various RGB objects, we retrained the tracking CNN and modified the binarization (which was very specific for ships) allows for its application across various categories, including cars, dogs, deer, giraffes, and many other.

\subsection{Infrared Ship Segmentation}
\label{ISSaOC}
\begin{figure*}
\centering
\subfloat[RGB-segm.] {\includegraphics[width=1.175in]{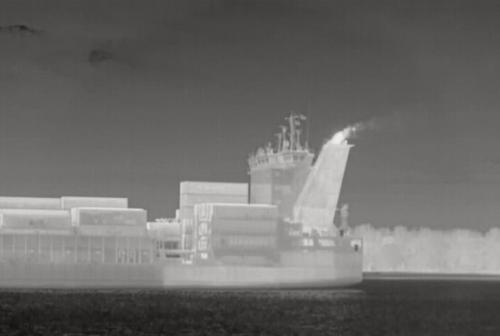}%
\label{image}}
\hfil
\subfloat[+ strategies]{\includegraphics[width=1.175in]{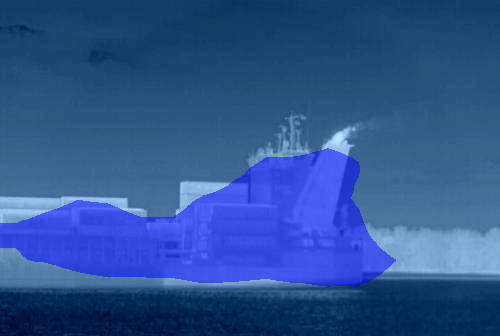}
\label{blueimage}}
\hfil
\subfloat[+ CRF]{\includegraphics[width=1.175in]{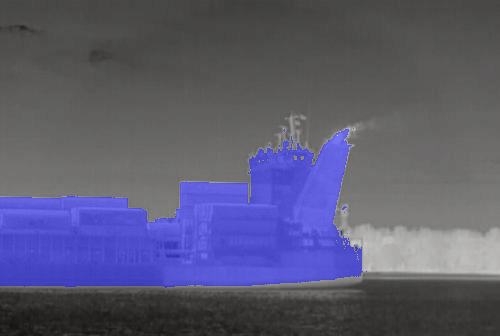}%
\label{segmcrf}}
\hfil
\subfloat[contour label]{\includegraphics[width=1.175in]{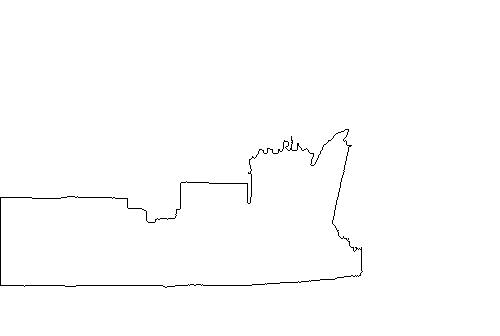}%
\label{contour}}
\caption{Visualization of \protect\subref{image} RGB segmentation method (no segment visible, like original image) \protect\subref{blueimage} + strategies, \protect\subref{segmcrf} + CRF, and \protect\subref{contour} conversion to contour label; images from \cite{Kelm2025}.}
\label{infrasegmzugang}
\end{figure*}

Fig.~\ref{infrasegmzugang} illustrates the access to the ship's IR object contour detection.
By using two strategies: a highly unbalanced RGB ship dataset, so that other classes are still considered, but the algorithm's response to ship segmentation is overly emphasized, and an adjustment of the IR image intensity channel to better match the typical color channel values of ship images, so that the RGB segmentation becomes more sensitive to IR.
A Conditional Random Field (CRF) method \cite{crf} occasionally creates meaningful labels from it.
Self-training generates an appropriate number of labels. It was stopped at 227 images.
Resulting masks are then converted to train an object contour detector for IR data.
Specifically, by adapting the RCN\cite{rcn} model to produce weights for RCN-IR\footnote{\url{https://github.com/AndreKelm/RefineContourNet/tree/master/refinenet-contour-master/model\_trained}}.

\subsection{Segmentation of Various Objects}
\label{CBSoO}
To extend WtL to more objects than only ship, we selected 90 COCO \textit{test-dev} \cite{COCO} images with a typical object focus and manually refined the rough annotations in detail. 
We call this small subset of COCO: DOC (Detailed Object Contour) dataset and publish these together with our 50 validation images and their ground truth.
For details on training the flat tracking CNN and its architecture, see the WtL paper \cite{9412410}.

The WtL binarization method was developed originally for ships, where the longest line from the soft contour map is used to create a starting condition for thresholding, which searches for the lowest value at which the object contour is closed. Thresholding is done by artificially splitting the object contour by a separation line and identifying two neighboring points that, when reconnected, indicate that the object contour is closed.
Objects other than ships usually do not have a typical waterline visible from the side.
WtL2 uses the same thresholding method as WtL, but starts with a more general assumption for different types of objects.
Algorithm \ref{ObjContConmple} takes the WtL contour, a grayscale image, as input and produces a binary WtL2 output.
\begin{algorithm}[b!]
 \caption{\textit{object contour binarization}}
 \begin{algorithmic}[1]
 \renewcommand{\algorithmicrequire}{\textbf{input:}}
 \renewcommand{\algorithmicensure}{\textbf{output:}}
 \REQUIRE WtL contour
 \ENSURE binary WtL2
\\ \textit{initialization}: create list $L(p,x)$ of pairs (pixel, value) from WtL contour
\REPEAT
   \STATE $p_{x}$ $\leftarrow$ max($x$) in $L(p,x)$
    \STATE edgegradient($p_{x}$) $\leftarrow$ sobel-filter(gaussian-filter(WtL contour($p_{x}$)))
  \STATE $separationline$ $\leftarrow$ edgegradient($p_{x}$)
  \STATE optimal $pixel_{1}$, $pixel_{2}$ $\leftarrow$ $separationline$ 
  \STATE $2\:endpoints$ $\leftarrow$ $separationline$ 
    \IF {$endpoints$ connect object and background}
  \STATE save $p_{x}$
    \ELSE 
    \STATE delete $p_{x}$ from L
  \ENDIF
\UNTIL{$p_{x}$ is not deleted}
\STATE binary WtL2 $\leftarrow$ \textbf{thresholding}($pixel_{1}$, $pixel_{2}$, $separationline$)
\RETURN binary WtL2
 \end{algorithmic}
 \label{ObjContConmple}
\end{algorithm}
Initialization includes a list of likely object contour pixels.
In code, a loop checks the remaining highest pixel intensity to see if its location is suitable for the separation line. 
The condition is met if the separation line points orthogonally from the background to the object contour or vice versa, and if the end points of the separation line touch the two largest enclosed areas, which are expected to be the background and the largest area in the object.
If not, the pixel is discarded and the next one is evaluated. Once a suitable pixel and corresponding separation line are identified, WtL thresholding searches for an optimal value to close the object contour and returns a binary image (see \cite{9412410} for details).
\section{Evaluation}
We first compare the two WtL versions using our 50 DOC-val dataset (section\,\ref{CBSoO}) to show the progress made. 
For IR ship segmentation, we could not find any public data/methods for a comparable evaluation, 
so we are releasing complementary data to MASSmind, the Elbe Ship IR and RGB Image Dataset
(ESIRRID\footnote{https://cloud.uni-hamburg.de/s/YKN8Lqe58tS2TdR}), which consists of unregistered RGB and IR images captured in the Lower Elbe River region, along with our self-annotated 10 Detailed IR Ship Contour (DIRSC) Ground Truth (GT); available in the WtL2 GitHub repository to enable comparisons.
To investigate the performance of the innovative contour tracking approach in WtL2 we build our own baseline using RefineNet (RN) \cite{lin2019refinenet}, which we call RN-IR. We used 227 labels generated by self-training (section\,\ref{ISSaOC}) for its supervised fine-tuning. RN-IR uses a ResNet101 backbone with training, similar to the object contour detector, allowing a direct comparison between conventional DL and the innovative WtL2.
When we evaluate objects in color, we follow a similar baseline. There are not many methods similar to our approach, so we focused on the sota contour-based segmentation method E2EC.

\subsection{WtL vs. WtL2}
We compare the two WtL versions using our 50 DOC-val dataset (see section\,\ref{CBSoO}) to show the progress made, which is clearly visible in the Table \ref{table_closed_contours_IoU} for all metrics. 
\begin{table*}[b!]
\renewcommand{\arraystretch}{1.15}
\setlength{\tabcolsep}{3pt} 
\caption{Comparison of NMS, WtL, and WtL2 algorithms on 50 DOC validation images, evaluating closed shapes by absolute number, percentage, and IoU.}
\label{table_closed_contours_IoU}
\centering
\small 
\begin{tabular}{|c|c|c|c|}
\hline
method & closed shapes & percentage of closed shape & IoU \\ \hline
\hline
NMS & 2 & 4\% & - \\ \hline
WTL & 28 & 56\% & 44.57 \\ \hline
WTL2 & \textbf{40} & \textbf{80\%} & \textbf{75.74} \\ \hline
\end{tabular}
\end{table*}
WtL2 creates much more closed shapes of objects, resulting in a higher IoU when these shapes are transformed into masks and used for segmentation.
\subsection{Infrared Ship Segmentation}
Some methods focus on their specific and non-public test data and do not always provide model weights \cite{compare,noinfosshipsegm}. 
So we manually label our own test data, 10 Detailed InfraRed Ship Contour (DIRSC) images,
along with our very robust baseline\footnote{https://github.com/AndreKelm/Infrared-Ship-and-RGB-Ship-Scene-Segmentation}
, both of which we provide for comparison.
\begin{table*}
\renewcommand{\arraystretch}{1.150} 
\setlength{\tabcolsep}{3pt} 
\caption{IR ship segmentation results on 10 DIRSC validation images, sorted by highest IoU. NMS and binary WtL2 are evaluated for closed object contours. If a closed contour is detected, the row is marked in gray.}
\label{table_results_DIRSC}
\centering
\small
\resizebox{\textwidth}{!}{
\begin{tabular}{|c|c|c|c|c|c|c|c|c|}
\hline
method & NMS & binary WtL2 & \multicolumn{3}{c|}{RN-IR} & \multicolumn{3}{c|}{WtL2-seg-IR} \\ \hline
image & closed shape & closed shape & P & R & IoU & P & R & IoU \\ \hline
\hline
\rowcolor{gray!32} 1 & N/A & \textbf{yes} & \textbf{98,00} & 95,74 & 93,90 & 97,24 & \textbf{98,56} & \textbf{95,88} \\ \hline
\rowcolor{gray!32} 2 & N/A & \textbf{yes} & 96,09 & \textbf{96,39} & \textbf{92,75} & \textbf{96,44} & 95,84 & 92,57 \\ \hline
\rowcolor{gray!32} 3 & N/A & \textbf{yes} & 94,15 & \textbf{96,13} & 90,71 & \textbf{96,11} & 96,00 & \textbf{92,41} \\ \hline
\rowcolor{gray!32} 4 & N/A & \textbf{yes} & 92,90 & 96,24 & 89,65 & \textbf{93,27} & \textbf{97,21} & \textbf{90,83} \\ \hline
\rowcolor{gray!32} 5 & N/A & \textbf{yes} & \textbf{90,57} & 97,95 & \textbf{88,89} & 87,93 & \textbf{98,73} & 86,95 \\ \hline
6 & N/A & N/A & \textbf{82,38} & \textbf{99,04} & \textbf{81,73} & 0 & 0 & 0\\ \hline
7 & N/A & N/A & 98,19 & \textbf{78,97} & \textbf{77,84} & \textbf{100} & 0,05 & 0,05 \\ \hline
8 & N/A & N/A & \textbf{71,68} & \textbf{90,49} & \textbf{66,67} & 8,89  &  0,02 & 0,02 \\ \hline
\rowcolor{gray!32} 9 & N/A & \textbf{yes} & \textbf{66,69} & 93,48 & \textbf{63,69} & 63,58 & \textbf{94,72} & 61,40 \\ \hline
10 & N/A & N/A & \textbf{23,73} & \textbf{80,96} & \textbf{22,47} & 0,05 & 0 & 0 \\ \hline
\hline
$\varnothing$ & 0\,\% & \textbf{60\,\%} & \textbf{81,44} & \textbf{92,54} & \textbf{76,83} & 69,35 & 58,11 & 52,01 \\ \hline
\rowcolor{gray!32} $\varnothing_{1, 2, 3, 4, 5, 9}$ &  0\,\% & \textbf{100\,\%} & \textbf{89,73} & 95,99 & 86,60 & 89,10 & \textbf{96,84} & \textbf{86,67} \\ \hline
\end{tabular}
}
\end{table*}
Table \ref{table_results_DIRSC} shows all 10 val images. In terms of total IoU, WtL2-seg-IR struggles against our robust baseline, which is 
already at a high level with an average of about 77 IoU. 
The extremes of the algorithm become visible. Sometimes there are complete failures in IoU, but then there are exceptionally high peaks. Noteworthy is the high recall, a property of WtLs. 
Focusing only on the images where the algorithm works as intended (marked in gray), it slightly outperforms the comparable baseline.
It is noteworthy that only the object contour detector was adapted to IR, not the tracking CNN itself, which indicates a certain robustness of the core algorithm. 
Considering that our baseline already achieves an average value of 77 IoU, this performance by WtL2-seg-IR (with closed object contours) is remarkable.
\begin{figure}
    \centering
    \begin{minipage}{\textwidth}
        \raggedright
        \begin{minipage}[c]{0.04\textwidth} 
            (a)
        \end{minipage}%
        \begin{minipage}[c]{0.945\textwidth} 
            \includegraphics[valign=c,width=0.992\textwidth,]{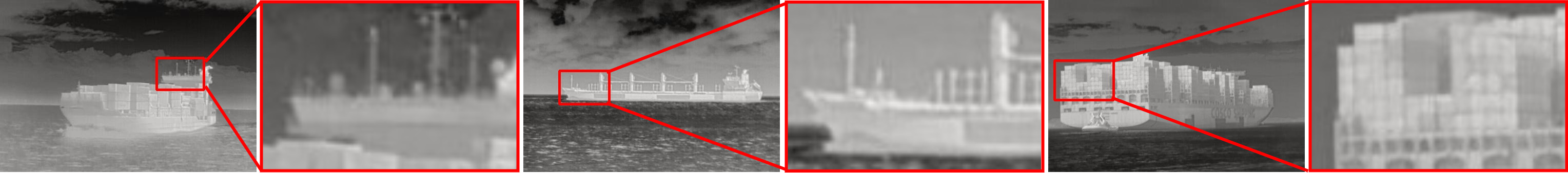}
        \end{minipage}
    \end{minipage}
    \vspace{0.0025\textwidth}
    \begin{minipage}{\textwidth}
        \raggedright
        \begin{minipage}[c]{0.04\textwidth}
            (b)
        \end{minipage}%
        \begin{minipage}[c]{0.945\textwidth}
            \includegraphics[valign=c,width=0.992\textwidth]{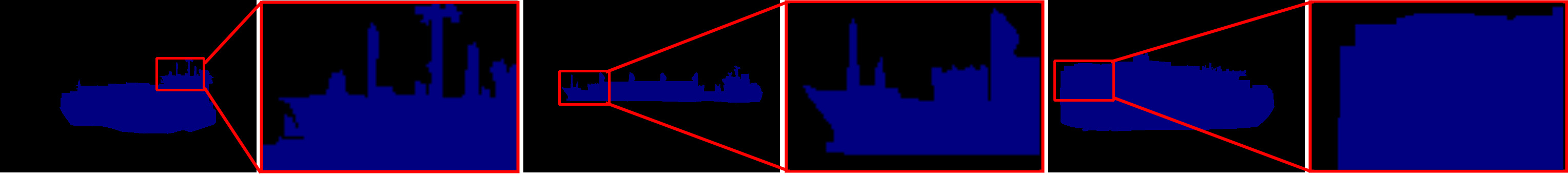}
        \end{minipage}
    \end{minipage}
    \vspace{0.0025\textwidth}
    \begin{minipage}{\textwidth}
        \raggedright
        \begin{minipage}[c]{0.04\textwidth}
            (c)
        \end{minipage}%
        \begin{minipage}[c]{0.945\textwidth}
            \includegraphics[valign=c,width=0.992\textwidth]{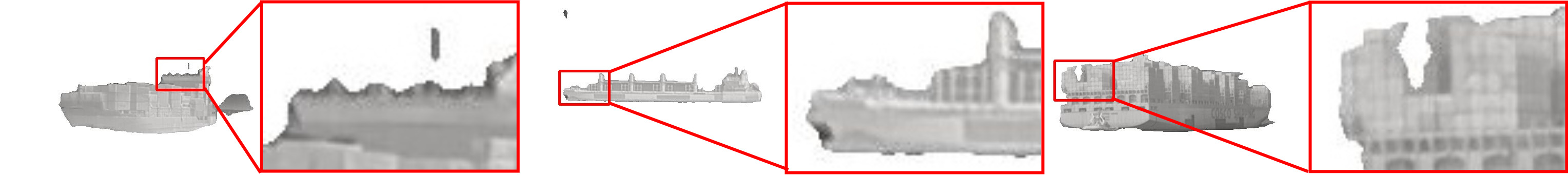}
        \end{minipage}
    \end{minipage}
    \vspace{0.0025\textwidth}
    \begin{minipage}{\textwidth}
        \raggedright
        \begin{minipage}[c]{0.04\textwidth}
            (d)
        \end{minipage}%
        \begin{minipage}[c]{0.945\textwidth}
            \includegraphics[valign=c,width=0.992\textwidth]{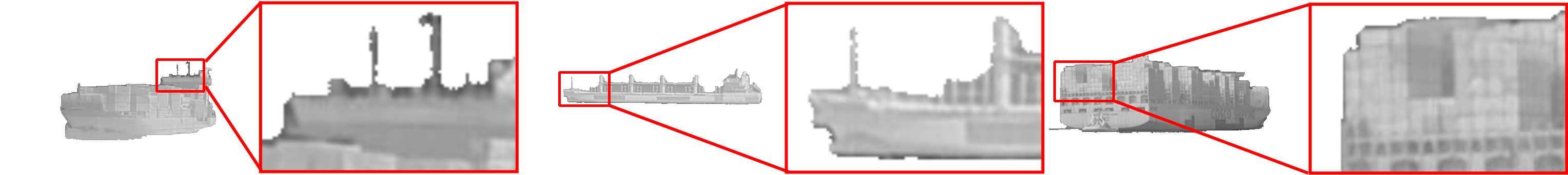}
        \end{minipage}
    \end{minipage}
    \vspace{0.0025\textwidth}
    \caption{Visualization of different IR ship segmentation results: (a) Original, (b) Ground Truth, (c) RefineNet-IR (baseline), (d) WtL2-seg-IR (ours); images from \cite{Kelm2025}.}
    \label{resultsobjectsWtL2IR}
\end{figure}
The Fig.\,\ref{resultsobjectsWtL2IR} shows no images where no object contour was closed and therefore no area for segmentation was found. Since these cases are clearly evident from the Table \ref{table_results_DIRSC} with very poor IoU. 
Fig.\,\ref{resultsobjectsWtL2IR} shows three different IR ships in comparison to our baseline.
For the ship on the left, the WtL2-seg-IR does not have a false positive area behind the ship, as the baseline does.
WtL2 incorrectly segments the bow wave, but performed significantly better in detecting the antennas.
The middle image shows a similar segmentation for both in the first glance, but a better detailed bow (including anchor, stem, and mast) is achieved with WtL2-seg-IR.
On the right, the stern of the large, angular container ship is much better segmented. Even the bridge wing is accurately extracted.
\subsection{Detailed Segmentation of Various Objects}
\begin{figure}[t!]
    \centering
    \begin{minipage}{\textwidth}
        \raggedright
        \begin{minipage}[c]{0.04\textwidth} 
            (a)
        \end{minipage}%
        \begin{minipage}[c]{0.945\textwidth} 
        \includegraphics[valign=c,width=0.99\textwidth,]{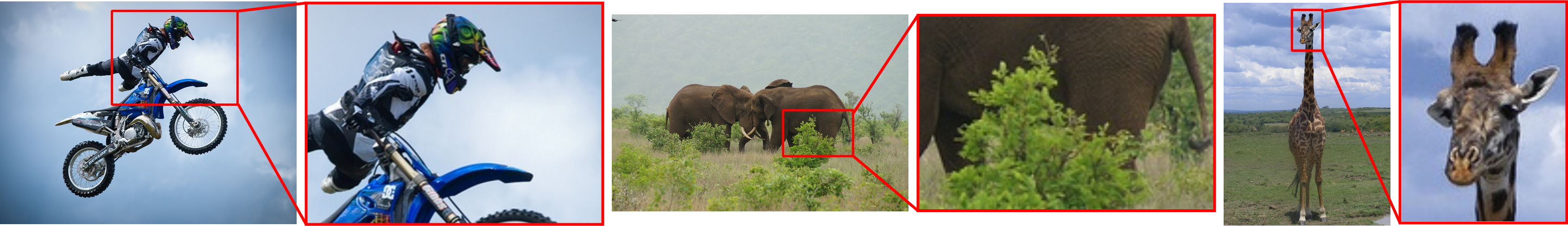}
        \end{minipage}
    \end{minipage}
    \vspace{0.0025\textwidth}
    \begin{minipage}{\textwidth}
        \raggedright
        \begin{minipage}[c]{0.04\textwidth}
            (b)
        \end{minipage}%
        \begin{minipage}[c]{0.945\textwidth}    \includegraphics[valign=c,width=0.99\textwidth]{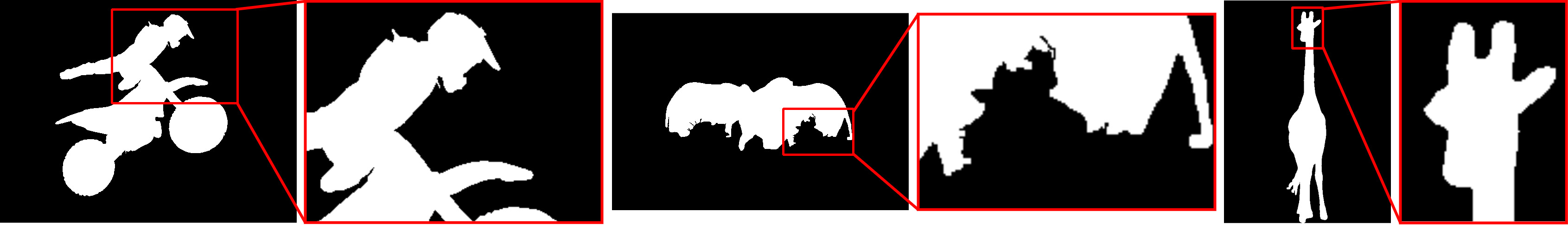}
        \end{minipage}
    \end{minipage}
    \vspace{0.0025\textwidth}
    \begin{minipage}{\textwidth}
        \raggedright
        \begin{minipage}[c]{0.04\textwidth}
            (c)
        \end{minipage}%
        \begin{minipage}[c]{0.945\textwidth}    \includegraphics[valign=c,width=0.99\textwidth]{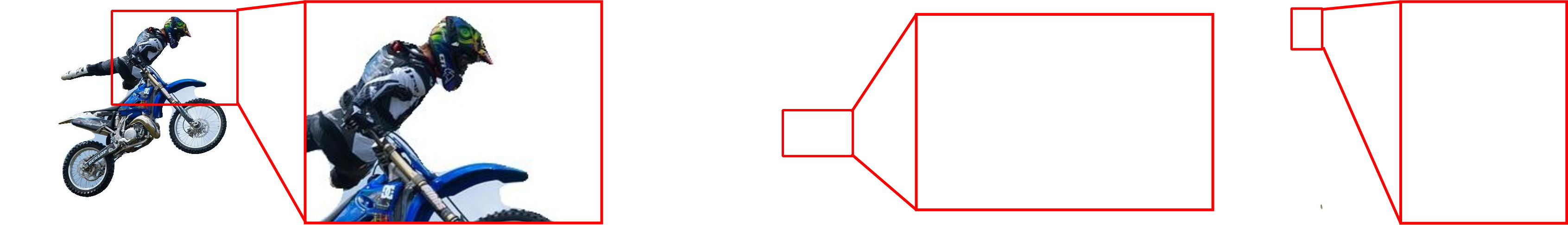}
        \end{minipage}
    \end{minipage}
    \vspace{0.0025\textwidth}
    \begin{minipage}{\textwidth}
        \raggedright
        \begin{minipage}[c]{0.04\textwidth}
            (d)
        \end{minipage}%
        \begin{minipage}[c]{0.945\textwidth}
\includegraphics[valign=c,width=0.99\textwidth]{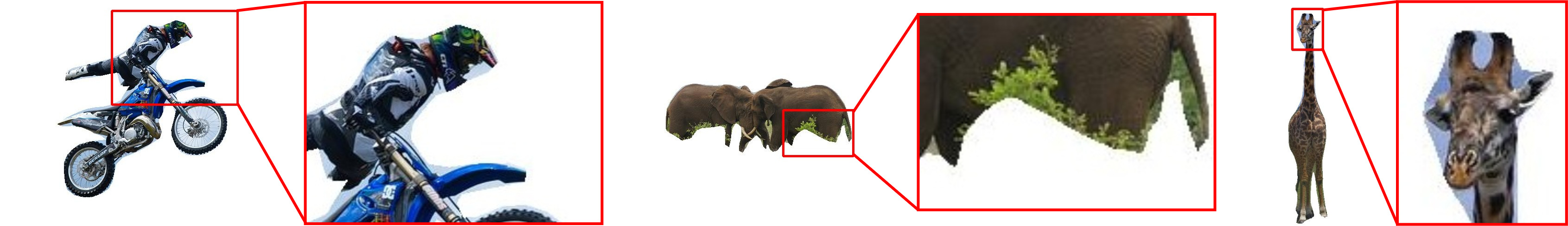}
        \end{minipage}
    \end{minipage}
    \vspace{0.0025\textwidth}
    \begin{minipage}{\textwidth}
        \raggedright
        \begin{minipage}[c]{0.04\textwidth}
            (e)
        \end{minipage}%
        \begin{minipage}[c]{0.945\textwidth}
            \includegraphics[valign=c,width=0.99\textwidth]{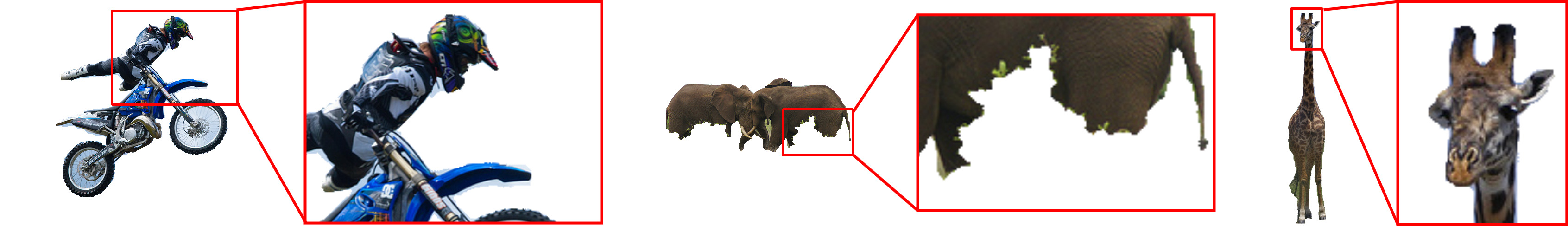}
        \end{minipage}
    \end{minipage}
    \vspace{0.0025\textwidth}
    \caption{Visualization of different  segmentation results: (a) Original, (b) Ground Truth, (c) RefineNet (baseline) \cite{lin2019refinenet}, contour-based segmentation: (d) E2EC (sota) and
    (e) WtL2-seg (ours); images from COCO \cite{COCO}.}
    \label{resultsobjectsWtL2rgb}
\end{figure}

While recent contour-based segmentation methods are evaluated on COCO, our objective differs. 
We aim for highly detailed segmentations, which COCO does not provide. Therefore, we updated 50 COCO \textit{test-dev} labels to align with our needs.
The first row in Table \ref{table_results_DOC} reveals the disadvantage of the method's robustness, since a further examination shows that for some images there is almost a total failure in terms of IoU.
\begin{table*}[b!]
\renewcommand{\arraystretch}{1.15}
\setlength{\tabcolsep}{3pt} 
\caption{Object segmentation results for 50 refined COCO \textit{test-dev} for our baseline RN and two contour-based methods E2EC and WtL2-seg. The top row shows the overall result for all images. The second row (marked in gray) shows the 40 images for which WtL2 was able to form a closed object contour.}
\label{table_results_DOC}
\centering
\small
\begin{tabular}{|c|c|c|c|c|c|c|c|c|c|}
\hline
method & \multicolumn{3}{c|}{RN} & \multicolumn{3}{c|}{E2EC} & \multicolumn{3}{c|}{WtL2-seg} \\ \hline
metric & P & R & IoU & P & R & IoU & P & R & IoU \\ \hline
\hline
$\varnothing_{{\text{50: all images}}}$ & \textbf{93.14} & 79.02 & 75.04 & 89.31 & \textbf{89.71} & \textbf{84.50} & 89.55 & 81.02 & 75.74\\ \hline
\rowcolor{gray!32} $\varnothing_{{\text{40: cl. shapes}}}$ & 92.09 & 80.36 & 75.95 & 89.40 & 89.84 & 84.06 & \textbf{93.58} & \textbf{94.83} & \textbf{89.01}\\ \hline
\end{tabular}
\end{table*}
However, the second row shows the advantage when it works and a closed object contour is formed: a high level of detail is given. In this case, the IoU even outperforms the sota in contour-based approaches.
For illustration, Fig.\,\ref{resultsobjectsWtL2rgb} shows images from this subset. WtL2 extracts significantly more detail than the baseline and E2EC overall. This includes features such as the motorcyclist's helmet shield and tire treads, the elephant's belly and right tail, and the giraffe's hump, ears, mouth, neck, tail and legs.
\section{Conclusion}
We have enhanced a unique contour tracking method for contour refinement. To demonstrate the potential of this innovative approach, we adapted it for the niche of IR ship segmentation and generally extended it to various objects beyond ships in the RGB domain.
In typical foreground-background scenarios, it outperforms the sota contour-based methods when forming a closed object contour, and it segments common categories such as dogs and cars, as well as rare and unusual categories such as deer, giraffes, and elephants, in a robust and detailed manner.
This makes it a compelling method for specialized applications 
requiring detailed segmentation or the production of high-quality samples,
potentially accelerating development in niche areas of computer vision, even with complex object contours or masks.
%
%
%
\bibliographystyle{splncs04}
\bibliography{mybibliography}

\end{document}